\documentclass{article}
\usepackage{spconf,amsmath,graphicx}
\usepackage{amsfonts}
\usepackage{algorithmic}
\usepackage{algorithm}
\usepackage{array}
\usepackage[caption=false,font=normalsize,labelfont=sf,textfont=sf]{subfig}
\usepackage{textcomp}
\usepackage{stfloats}
\usepackage{url}
\usepackage{verbatim}
\usepackage{graphicx}
\usepackage{multirow}
\usepackage[table,xcdraw]{xcolor}
\usepackage[sort&compress,numbers,square]{natbib}
\usepackage{times}
\usepackage{epsfig}
\usepackage{graphicx}
\usepackage{amsmath}
\usepackage{amssymb}
\usepackage{booktabs}
\usepackage{amsmath,amssymb,amsfonts}
\usepackage{algorithmic}
\usepackage{textcomp}
\usepackage{xcolor}
\usepackage{multirow}
\usepackage[sort&compress,numbers,square]{natbib}
\usepackage{stfloats}
\usepackage{enumitem}
\usepackage{soul}
\usepackage{hhline}

\title{On Pixel-level Performance Assessment in Anomaly Detection}
\name{Mehdi Rafiei\textsuperscript{1}, Toby P. Breckon\textsuperscript{2}, and Alexandros Iosifidis\textsuperscript{1}\vspace{-0.2cm}}
\address{\textsuperscript{1}\textit{DIGIT, Department of Electrical and Computer Engineering,}
\textit{Aarhus University, Aarhus, Denmark} \\ \textsuperscript{2}\textit{Department of Computer Science,}\textit{Durham University, Durham, UK} \\
rafiei@ece.au.dk, toby.breckon@durham.ac.uk, ai@ece.au.dk\vspace{-0.1cm}}
\begin{document}
\topmargin=0mm
%\ninept
%
\maketitle

\begin{abstract}
Anomaly detection methods have demonstrated remarkable success across various applications. However, assessing their performance, particularly at the pixel-level, presents a complex challenge due to the severe imbalance that is most commonly present between normal and abnormal samples. Commonly adopted evaluation metrics designed for pixel-level detection may not effectively capture the nuanced performance variations arising from this class imbalance. In this paper, we dissect the intricacies of this challenge, underscored by visual evidence and statistical analysis, leading to delve into the need for evaluation metrics that account for the imbalance. We offer insights into more accurate metrics, using eleven leading contemporary anomaly detection methods on twenty-one anomaly detection problems. Overall, from this extensive experimental evaluation, we can conclude that Precision-Recall-based metrics can better capture relative method performance, making them more suitable for the task.
\end{abstract}
\begin{keywords}
anomaly detection, class imbalance, performance metric selection.
\end{keywords}\vspace{-0.2cm}
\section{Introduction}\vspace{-0.2cm}
Anomaly detection plays an important role across various applications, including security, product inspection, and medical diagnostics. Contemporary methods have demonstrated remarkable success by achieving high performance levels, primarily through widely used image and pixel-level metrics. However, an underlying conundrum emerges due to the inherent imbalance between normal and abnormal pixels that commonly characterize anomaly detection problems. This intricacy in dataset composition poses a critical challenge to the practical evaluation of model performance.

Current approaches often conceptualize visual anomaly detection as a high-level per-image binary decision task or a low-level per-pixel segmentation task. Existing literature relies on performance metrics such as the Area Under the Receiver Operating Characteristic Curve (AUROC) and the F1-score to assess image-level performance. Similarly, metrics like AUROC and the Area Under the Per-Region-Overlap curve (AUPRO) \citep{gudovskiy2022cflow} gauge pixel-level performance of anomaly detection models. While these metrics are very well suited and effective for image-level anomaly detection, their utility diminishes when employed for pixel-level performance evaluation due to the substantial class imbalance between abnormal and normal pixels in any given sample. 

In this paper, to statistically and visually illustrate the severe imbalance between normal and abnormal classes appearing in commonly adopted benchmarks for pixel-level anomaly detection, we use five datasets containing 21 separate anomaly detection problems \citep{bergmann2019mvtec, mishra2021vt, bovzivc2021mixed, krohling2019bracol, rafiei2022recognition}. We introduce the commonly adopted pixel-level metrics, their limitations, and alternative Precision-Recall-based metrics that are more suitable for pixel-level anomaly detection. Using eleven leading contemporary anomaly detection methods \citep{lee2022cfa, gudovskiy2022cflow, shi2022csflow, ahuja2019probabilistic, zavrtanik2021draem, batzner2023efficientad, yu2021fastflow, defard2021padim, roth2022towards, deng2022anomaly, wang2021student}, we provide comprehensive experimental evidence that the current practice yields closely clustered performance values across different methods, making it challenging to discern subtle performance differences of anomaly detection methods. Finally, we show the effectiveness of the Precision-Recall-based metrics in better capturing relative method performance, making them more suitable for performance comparison of pixel-level anomaly detection tasks.\vspace{-0.2cm}

% In this paper, we introduce this performance metric selection challenge and provide experimental evidence that the current practice employing \hl{AAA}. We provide a comprehensive performance evaluation of \hl{BBB}. In this regard:
% \begin{itemize}[leftmargin=*]
%     \vspace{-0.2cm}
%     \item five datasets containing 21 separate anomaly detection problems are used to statistically and visually illustrate the severe imbalance between normal and abnormal samples, 
%     \vspace{-0.3cm}
%     \item the commonly adopted pixel-level metrics, their limitations, and the alternative metrics are described, 
%     \vspace{-0.3cm}
%     \item the image and pixel-level performance of eleven leading contemporary anomaly detection methods on all datasets are reported to demonstrate the impact of imbalanced datasets on different metrics.
% \end{itemize}

\section{Visual and Statistical Data Analysis}\vspace{-0.2cm}
To underline the significance of this challenge, we provide visual and statistical analysis of the pixel-level class imbalance within five datasets:
\begin{itemize}[leftmargin=*]
    \vspace{-0.2cm}
    \item \textbf{MVTec AD} \citep{bergmann2019mvtec} contains 5,354 industrial samples divided into 15 classes (i.e., Bottle, Cable, Capsule, Carpet, Grid, Hazelnut, Leather, Metal nut, Pill, Screw, Tile, Toothbrush, Transistor, Wood, and Zipper). 
    \vspace{-0.3cm}
    \item \textbf{BeanTech AD} \citep{mishra2021vt} contains 2,542 real-world industrial samples, including body and surface defects, divided into three classes. 
    \vspace{-0.3cm}
    \item \textbf{KSDD2} \citep{bovzivc2021mixed} was created for detecting surface defects. The dataset comprises a training set containing 246 abnormal and 2,085 normal samples targeting supervised, weakly supervised, or unsupervised anomaly detection. We used only normal specimens for training.
    \vspace{-0.3cm}
    \item \textbf{Coffee leaf diseases AD} \citep{krohling2019bracol} contains 1,664 samples, divided into 400 healthy and 1,264 sick leaves. This dataset is designed for the semantic segmentation task of classifying image pixels to leaf, diseases (i.e., miner, rust, and phoma), and background. We modified the annotation masks for anomaly detection by merging all disease labels into one anomaly label.
    \vspace{-0.3cm}
    \item \textbf{X-ray Fibrus Product AD} \citep{rafiei2022recognition} contains 478 X-ray samples of fibrous products initially developed for classification tasks. We added the corresponding segmentation masks to use the dataset for the anomaly detection task. 
\end{itemize}\vspace{-0.15cm}

\begin{figure}
\centering
\includegraphics[width=\columnwidth]{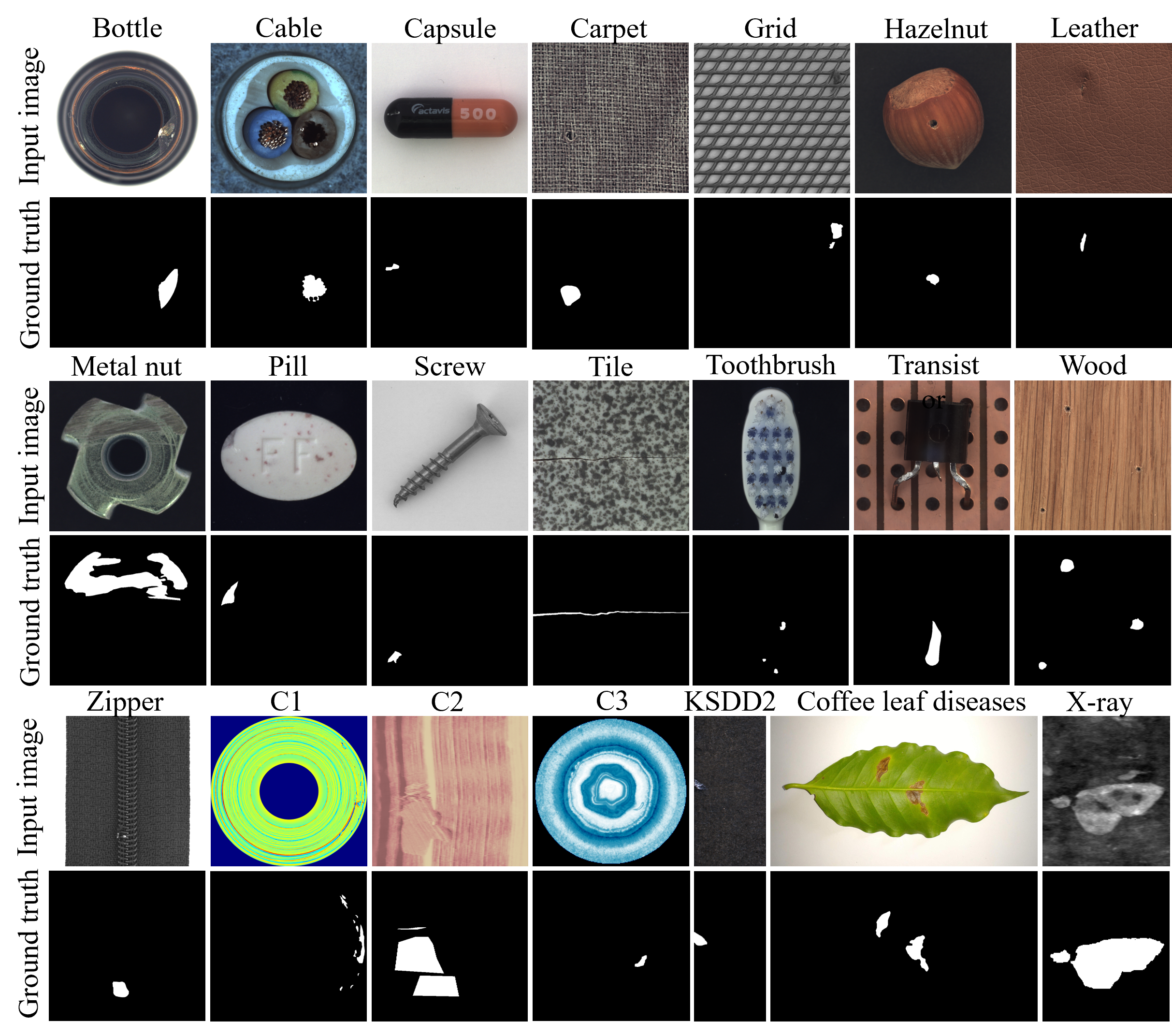}\vspace{-0.4cm}
\caption{Input image and ground truth samples of all MVTec AD classes, all BeanTech AD classes, KSDD2, Coffee leaf diseases AD, and X-ray Fibrus Product AD datasets.}\vspace{-0.4cm}
\label{data_samples}
\end{figure}
% \usepackage{multirow}

\begin{table*}[!b]
\vspace{-0.6cm}
\caption{\label{data} Statistical overview of the datasets.}\vspace{0.1cm}
\centering
\footnotesize
\begin{tabular}{ccccccccc}
\hline
\rowcolor[HTML]{9B9B9B} 
\multicolumn{2}{c}{\cellcolor[HTML]{9B9B9B}}  & \cellcolor[HTML]{9B9B9B}  & \multicolumn{3}{c}{\cellcolor[HTML]{9B9B9B}Test (Image-level)}   & \multicolumn{3}{c}{\cellcolor[HTML]{9B9B9B}Test (Pixel-level)}   \\ \cline{4-9} 
\rowcolor[HTML]{9B9B9B} 
\multicolumn{2}{c}{\multirow{-3}{*}{\cellcolor[HTML]{9B9B9B}Dataset / Class}} & \multirow{-3}{*}{\cellcolor[HTML]{9B9B9B}Train samples} & Normal & Abnormal & Ratio & Normal \%\ & Abnormal \% & Ratio \\ \hline
  & bottle  & 209 & 20  & 63  & 0.32  & 94.22  & 5.78 & 16.30 \\ \cline{2-9} 
  & cable & 224 & 58  & 92  & 0.63  & 97.13  & 2.87 & 33.84 \\ \cline{2-9} 
  & capsule & 219 & 23  & 109 & 0.21  & 99.08  & 0.92 & 107.70  \\ \cline{2-9} 
  & carpet  & 280 & 28  & 89  & 0.31  & 98.4 & 1.6  & 61.50 \\ \cline{2-9} 
  & grid  & 264 & 21  & 57  & 0.37  & 99.31  & 0.69 & 143.93  \\ \cline{2-9} 
  & hazelnut  & 391 & 40  & 70  & 0.57  & 97.87  & 2.13 & 45.95 \\ \cline{2-9} 
  & leather & 245 & 32  & 92  & 0.35  & 99.35  & 0.65 & 152.85  \\ \cline{2-9} 
  & metal\_nut  & 220 & 22  & 93  & 0.24  & 88.28  & 11.72  & 7.53  \\ \cline{2-9} 
  & pill  & 267 & 26  & 141 & 0.18  & 96.64  & 3.36 & 28.76 \\ \cline{2-9} 
  & screw & 320 & 41  & 119 & 0.34  & 99.75  & 0.25 & 399.00  \\ \cline{2-9} 
  & tile  & 230 & 33  & 84  & 0.39  & 92.96  & 7.04 & 13.20 \\ \cline{2-9} 
  & toothbrush  & 60  & 12  & 30  & 0.40  & 98.47  & 1.53 & 64.36 \\ \cline{2-9} 
  & transistor  & 213 & 60  & 40  & 1.50  & 96.95  & 3.05 & 31.79 \\ \cline{2-9} 
  & wood  & 247 & 19  & 60  & 0.32  & 93.93  & 6.07 & 15.47 \\ \cline{2-9} 
  & zipper  & 240 & 32  & 119 & 0.27  & 97.93  & 2.07 & 47.31 \\ \hhline{~*{8}{|-}|}
  & \cellcolor[HTML]{C0C0C0}Average & \cellcolor[HTML]{C0C0C0}242 & \cellcolor[HTML]{C0C0C0}31  & \cellcolor[HTML]{C0C0C0}84  & \cellcolor[HTML]{C0C0C0}0.37  & \cellcolor[HTML]{C0C0C0}96.68  & \cellcolor[HTML]{C0C0C0}3.32 & \cellcolor[HTML]{C0C0C0}29.12 \\ \hhline{~*{8}{|-}|} 
  & \cellcolor[HTML]{C0C0C0}Object classes average  & \cellcolor[HTML]{C0C0C0}236 & \cellcolor[HTML]{C0C0C0}33  & \cellcolor[HTML]{C0C0C0}88  & \cellcolor[HTML]{C0C0C0}0.38  & \cellcolor[HTML]{C0C0C0}96.63  & \cellcolor[HTML]{C0C0C0}3.37 & \cellcolor[HTML]{C0C0C0}28.67 \\ \hhline{~*{8}{|-}|} 
\multirow{-18}{*}{MVTec AD} & \cellcolor[HTML]{C0C0C0}Texture classes average & \cellcolor[HTML]{C0C0C0}253 & \cellcolor[HTML]{C0C0C0}27  & \cellcolor[HTML]{C0C0C0}76  & \cellcolor[HTML]{C0C0C0}0.36  & \cellcolor[HTML]{C0C0C0}96.79  & \cellcolor[HTML]{C0C0C0}3.21 & \cellcolor[HTML]{C0C0C0}30.15 \\ \hline
\rowcolor[HTML]{EFEFEF} 
\cellcolor[HTML]{EFEFEF}  & 1 & 400 & 21  & 50  & 0.42  & 96.74  & 3.26 & 29.67 \\ \hhline{~*{8}{|-}|} 
\rowcolor[HTML]{EFEFEF} 
\cellcolor[HTML]{EFEFEF}  & 2 & 400 & 30  & 200 & 0.15  & 94.53  & 5.47 & 17.28 \\ \hhline{~*{8}{|-}|}
\rowcolor[HTML]{EFEFEF} 
\cellcolor[HTML]{EFEFEF}  & 3 & 1000  & 400 & 41  & 9.76  & 99.79  & 0.21 & 475.19  \\ \hhline{~*{8}{|-}|} 
\rowcolor[HTML]{C0C0C0} 
\multirow{-4}{*}{\cellcolor[HTML]{EFEFEF}BeanTech AD} & Average & 400 & 150 & 93  & 1.61  & 97.02  & 2.98 & 32.56 \\ \hline
\multicolumn{2}{c}{KSDD2} & 2085  & 894 & 110 & 8.13  & 99.67  & 0.33 & 302.03  \\ \hline
\rowcolor[HTML]{EFEFEF} 
\multicolumn{2}{c}{\cellcolor[HTML]{EFEFEF}Coffee leaf diseases AD}  & 320 & 80  & 1264  & 0.06  & 98.51  & 1.49 & 66.11 \\ \hline
\multicolumn{2}{c}{X-ray} & 176 & 60  & 242 & 0.25  & 87.61  & 12.39  & 7.07  \\ \hline
\end{tabular}
\end{table*}

In Fig. \ref{data_samples}, we provide visual examples of the challenges associated with pixel-level class imbalance in the above-described datasets. This figure illustrates examples of input image samples and the related ground truth annotations from all classes, each corresponding to an anomaly detection problem. The white and black pixels in the annotations represent the abnormal and normal pixels, respectively. The severely lower number of white pixels in the annotations demonstrates the disparities between normal and abnormal pixel distributions, resulting in a substantial proportion of negative samples. 

To be more precise with respect to class imbalance ratios, Table \ref{data} outlines the characteristics of the datasets. The table showcases the ratio between normal and abnormal images and between normal and abnormal pixels in each class and the averages over each dataset. It can be seen that the normal-to-abnormal ratios on the image-level vary from 0.15 to 9.76 (except for the Coffee leaf disease AD dataset, which contains a high number of defective samples leading to a normal-to-abnormal ratio of 0.06). However, when considering pixel-level anomaly detection, the normal-to-abnormal ratios vary from 7.07 to 475.19, revealing the pronounced class imbalance at the pixel-level. The above underlines the complexity inherent in assessing the effectiveness of algorithms designed to detect abnormalities or anomalies at the pixel-level.

The inherent class imbalance in pixel-level anomaly detection brings forth a notable consequence: a trade-off between sensitivity and specificity emerges, which can diminish the responsiveness of commonly used performance metrics towards false negative or false positive classifications. Metrics like accuracy can offer misleading insights into model performance when the dataset comprises numerous negative samples by simply classifying most pixels as negative. Consequently, the occurrence of false negative pixels might not be adequately penalized using such metrics. Moreover, the abundance of negative pixels can significantly impact the ratio of false positives to true negatives, thereby reducing the precision of metrics in penalizing false positives. This situation creates a scenario where, while the model effectively detects anomalies, it lacks the precision to demarcate the defective regions. 
Hence, the intricacy of evaluating pixel-level performance originates from the complex interplay between class imbalance and the evaluation metrics employed. Addressing this challenge necessitates thoughtful contemplation of evaluation strategies tailored to the skewed class distribution. Techniques such as employing weighted metrics, utilizing resampling methods, or exploring alternative metrics that equally prioritize false positives and negatives can effectively capture the performance of a model in tasks involving imbalanced pixel-level data.\vspace{-0.2cm}

\section{Performance Metrics Assessment}
\begin{figure}
\centering
\includegraphics[width=0.9\columnwidth]{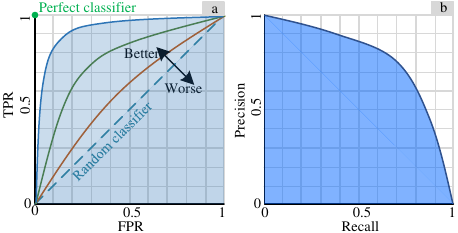}\vspace{-0.3cm}
\caption{a) Receiver Operator Characteristics (ROC) curve, b) Precision
vs. Recall curve. \citep{rafiei2023computer} }\vspace{-0.4cm}
\label{ROC}
\end{figure}

\begin{figure}
\centering
\includegraphics[width=\columnwidth]{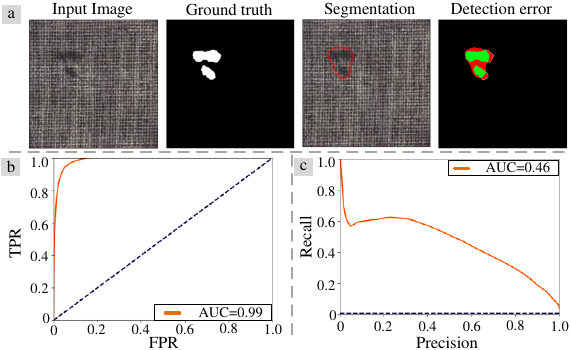}\vspace{-0.3cm}
\caption{Anomaly detection example using CFA \citep{lee2022cfa}: a) input image, ground truth, anomaly segmentation, and
detection error (red: false positive or false negative, green: true positive), b) ROC curve, and c) PR curve.}\vspace{-0.4cm}
\label{res_example}
\end{figure}

\vspace{-0.2cm}
The commonly adopted metrics for pixel-level anomaly detection are AUROC and AUPRO. As shown in Fig. \ref{ROC}a, AUROC is the area under the True Positive Ratio (TPR)-False Positive Ratio (FPR) curve. These two ratios are defined as $TPR = TP / (TP + FN)$ and $FPR = FP / (FP + TN)$, %:
where TP, TN, FP, and FN are the true positive, true negative, false positive, and false negative predictions. AUPRO is the modified AUROC that ensures both large and small anomalies are equally important in localization. AUROC and AUPRO are beneficial when dealing with binary classification problems where the distribution of classes is skewed, meaning one class has significantly fewer samples than the other. However, it can still be influenced by the overwhelming number of negative samples. This is because a method can achieve high TPR while FPR is still low due to the large number of negative pixels in the dataset. Fig. \ref{res_example}a shows an example of anomaly detection using CFA \citep{lee2022cfa}. The segmentation provided by the model almost covers the defective part but it also includes many non-defective pixels. Although the segmentation is not precise (it contains a considerable number of FP compared to TPs), the FPR value would be low due to the large number of negative pixels (high TN). Subsequently, while the TPR rises with different thresholds in the ROC curve, the FPR will stay low, causing a high AUROC value (Fig. \ref{res_example}-b). 

Precision-Recall-based metrics, such as the Area Under Precision-Recall curve (AUPR) and F1-score, tend to provide more meaningful performance assessment in such imbalanced scenarios. Precision has the same definition as TPR, measuring ``\textit{how many retrieved samples are relevant}," and Recall is defined as $Recall = TP / (TP + FN)$, measuring ``\textit{how many relevant samples are retrieved}." In other words, Precision focuses on the accuracy of positive classifications, while Recall focuses on how well the model identifies positive samples. These metrics are essential when the positive class (defective pixels) is the class of interest and one wants to know how well the model detects anomalies. 
As shown in Fig. \ref{ROC}b, AUPR is the area under the Precision-Recall curve. Therefore, in an anomaly detection case such as the one shown in Fig. \ref{res_example}a, although Recall is high ($\simeq1$), Precision can be low due to the number of FP. Fig. \ref{res_example}c shows the related PR curve and its higher ability to assess the method performance compared to the ROC curve (Fig. \ref{res_example}b).

As another Precision-Recall-based metric, the F1-score also considers both Precision and Recall, making it suitable for pixel-level anomaly detection with severe class imbalance.\vspace{-0.2cm}

\section{Visualization}\vspace{-0.2cm}
To visualize the suitability of the suggested metrics for pixel-level anomaly detection on highly class-imbalanced datasets, we report the performance achieved by eleven anomaly detection methods (i.e., CFA \citep{lee2022cfa}, CFLOW-AD \citep{gudovskiy2022cflow}, CSFlow \citep{shi2022csflow}, DFM \citep{ahuja2019probabilistic}, DRÆM \citep{zavrtanik2021draem}, EfficientAD \citep{batzner2023efficientad}, FastFlow \citep{yu2021fastflow}, PaDiM \citep{defard2021padim}, Patch-core \citep{roth2022towards}, Reverse Distillation \citep{deng2022anomaly}, and STFPM \citep{wang2021student}) on the five above-described datasets. In all cases, image-level performance is measured by using AUROC, AUPR, and F1-score, and pixel-level performance by using AUROC, AUPRO, AUPR, and F1-score. We use the prefixes I- and P- to indicate image-level and pixel-level performance, respectively.

\begin{figure*}
\centering
\includegraphics[width=\textwidth]{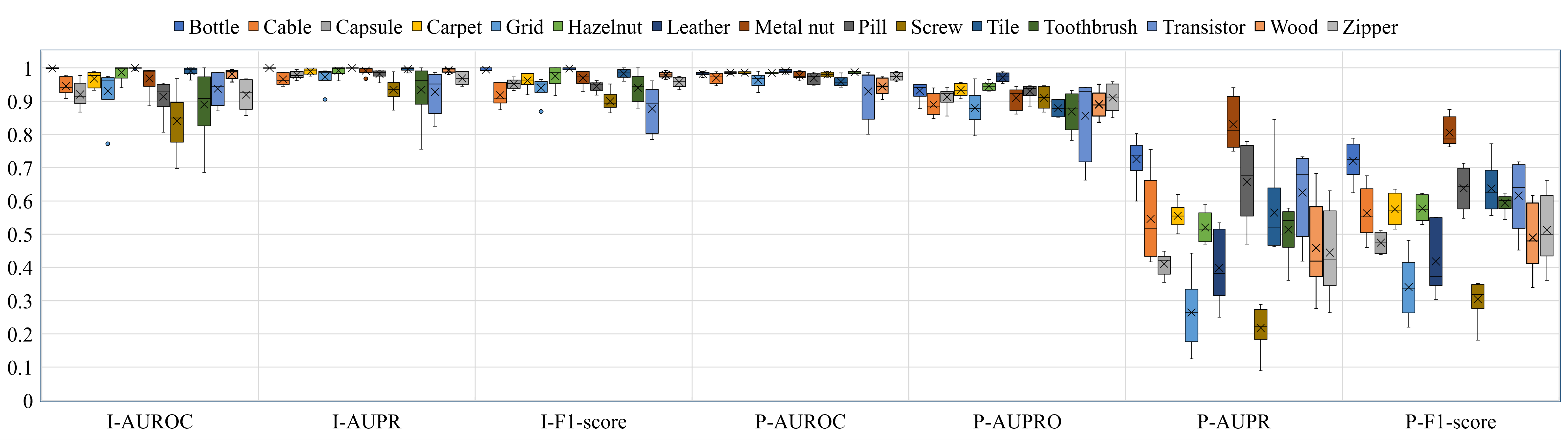}\vspace{-0.4cm}
\caption{The image and pixel-level performance metric ranges
achieved by the methods (all classes in MVTec AD dataset).}\vspace{-0.4cm}
\label{MVTec_all}
\end{figure*}

\begin{figure}
\centering
\includegraphics[width=\columnwidth]{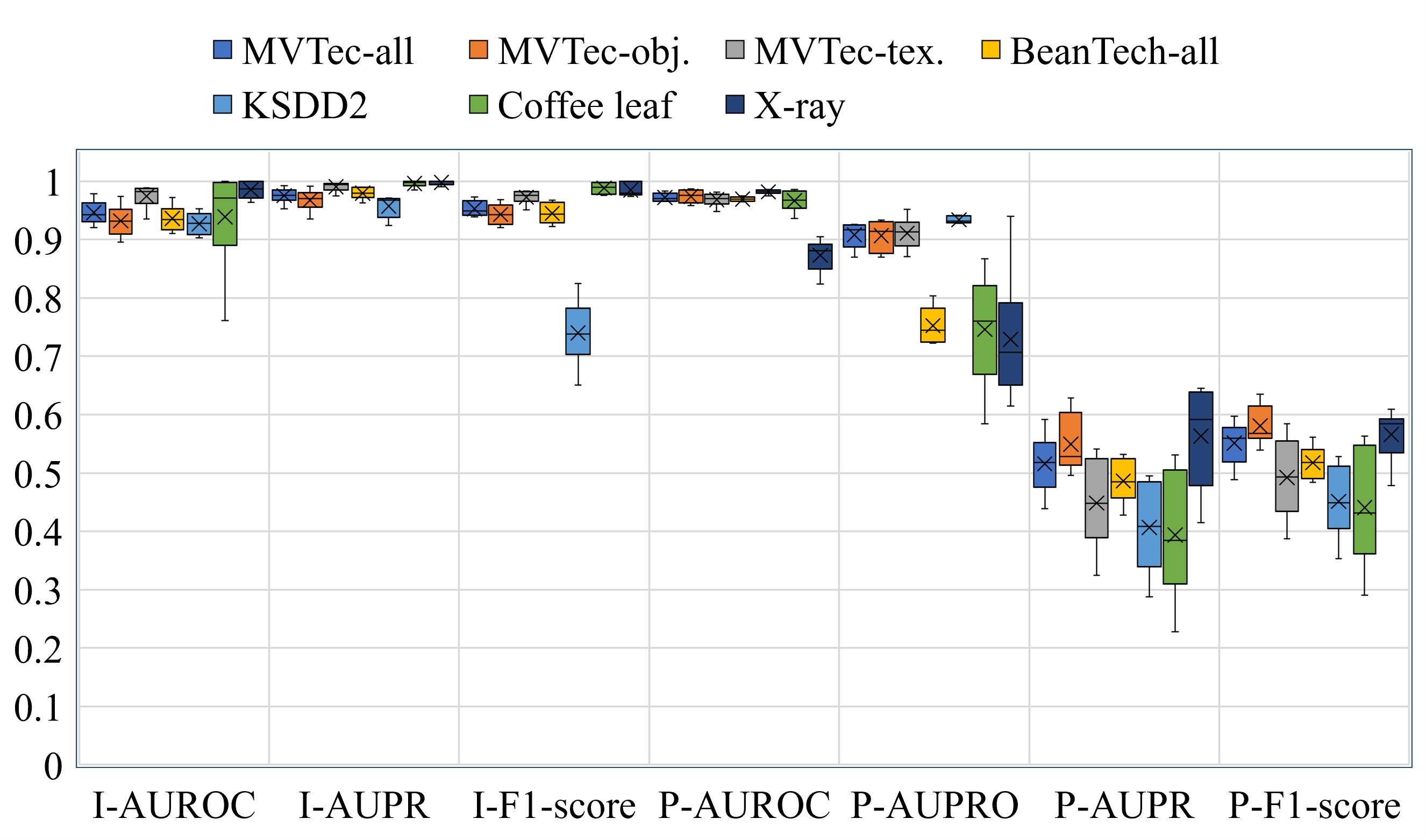}\vspace{-0.3cm}
\caption{The image and pixel-level performance metric ranges
achieved by the methods (averages on MVTec AD, overall average on BeanTech AD, KSDD2, Coffee leaf diseases AD, and X-ray Fibrus Product AD dataset).}\vspace{-0.4cm}
\label{All_res}
\end{figure}

Figs. \ref{MVTec_all} and \ref{All_res} illustrate the performance levels achieved by the methods on all 15 classes of MVTec AD and averages on each of the five datasets, respectively. Each box and whisker illustrate the performance range achieved by all anomaly detection methods for each performance metric.%
It can be seen that the use of the commonly adopted performance metrics for pixel-level anomaly detection (i.e., AUROC and AUPRO) leads to a high reported performance for all methods. Moreover, as denoted by the corresponding box sizes and whisker lengths, differences in performance are minor. On the other hand, the use of Precision-Recall-based metrics (i.e., AUPR and F1-score) leads to lower performance values and better highlights differences in the reported performances among the methods. In essence, employing metrics like AUROC and AUPR often yields closely clustered high-performance values across different methods that not only do not necessarily indicate accurate and precise pixel-level anomaly detection but also make it challenging to discern subtle differences in the performance of different methods. Conversely, utilizing metrics such as AUPR and F1-score introduces a broader spectrum of performance values due to their class imbalance considerations. Therefore, Precision-Recall-based metrics can better capture relative method performance, making them more suitable for performance comparison of anomaly detection methods.

Considering the performance levels reported in Figs. \ref{MVTec_all} and \ref{All_res} and the class ratios shown in Table \ref{data}, it can be seen that there is a direct relation between the level of class imbalance and the difference in the reported performance between the commonly adopted and Precision-Recall-based metrics. This direct relation indicates once more that the severe class imbalance is the reason behind the inability of the commonly adopted metrics to reflect the pixel-level performance of the methods precisely. This observation is further verified by considering that for image-level anomaly detection on the Coffee leaf disease AD dataset, F1-score is a more suitable metric as the dataset exhibits class imbalance on the image-level as well.\vspace{-0.2cm}

\section{Conclusion}\vspace{-0.2cm}
Evaluating pixel-level performance in anomaly detection presents a formidable challenge due to the prevalent class imbalance between normal and abnormal pixels. In this paper, we delved into the complexities of this challenge by employing visual evidence and statistical analysis. Additionally, we performed a comprehensive experimental study on the commonly adopted and suitable alternative metrics, encompassing various state-of-the-art anomaly detection methods and datasets, demonstrating the limitations of commonly adopted metrics when applied to pixel-level analysis. In conclusion, %considering the essential aspects of an anomaly detection task (e.g., detection accuracy, precise location segmentation, etc.), 
a thorough understanding of the pixel imbalance challenge, and suitable metric selection are crucial for developing robust evaluation strategies that enable accurate performance assessment of pixel-level anomaly detection methods. Overall, we concluded in this paper that Precision-Recall-based metrics are very well suited for pixel-level anomaly detection tasks, as they better capture relative method performance due to their class imbalance considerations.\vspace{-0.3cm}

\section*{Acknowledgment}\vspace{-0.2cm}
\noindent
The research leading to the results of this paper received funding from the Innovation Fund Denmark as part of MADE FAST.

\bibliographystyle{IEEEbib}
\bibliography{ref}

\begin{thebibliography}{10}

\bibitem{gudovskiy2022cflow}
Denis Gudovskiy, Shun Ishizaka, and Kazuki Kozuka,
\newblock ``Cflow-ad: Real-time unsupervised anomaly detection with
  localization via conditional normalizing flows,''
\newblock {\em IEEE Winter Conference on Applications of Computer Vision}, pp.
  98--107, 2022.

\bibitem{bergmann2019mvtec}
Paul Bergmann, Michael Fauser, David Sattlegger, and Carsten Steger,
\newblock ``Mvtec ad--a comprehensive real-world dataset for unsupervised
  anomaly detection,''
\newblock {\em IEEE Conference on Computer Vision and Pattern Recognition}, pp.
  9592--9600, 2019.

\bibitem{mishra2021vt}
Pankaj Mishra, Riccardo Verk, Daniele Fornasier, Claudio Piciarelli, and
  Gian~Luca Foresti,
\newblock ``Vt-adl: A vision transformer network for image anomaly detection
  and localization,''
\newblock {\em IEEE International Symposium on Industrial Electronics}, pp.
  01--06, 2021.

\bibitem{bovzivc2021mixed}
Jakob Bo{\v{z}}i{\v{c}}, Domen Tabernik, and Danijel Sko{\v{c}}aj,
\newblock ``Mixed supervision for surface-defect detection: From weakly to
  fully supervised learning,''
\newblock {\em Computers in Industry}, vol. 129, pp. 103459, 2021.

\bibitem{krohling2019bracol}
Renato~A Krohling, Jos{\'e} Esgario, and Jos{\'e}~A Ventura,
\newblock ``Bracol--a brazilian arabica coffee leaf images dataset to
  identification and quantification of coffee diseases and pests,''
\newblock {\em Mendeley Data}, vol. 1, 2019.

\bibitem{rafiei2022recognition}
Mehdi Rafiei, Dat~Thanh Tran, and Alexandros Iosifidis,
\newblock ``Recognition of defective mineral wool using pruned resnet models,''
\newblock {\em IEEE International Conference on Industrial Informatics}, pp.
  1--6, 2023.

\bibitem{lee2022cfa}
Sungwook Lee, Seunghyun Lee, and Byung~Cheol Song,
\newblock ``Cfa: Coupled-hypersphere-based feature adaptation for
  target-oriented anomaly localization,''
\newblock {\em IEEE Access}, vol. 10, pp. 78446--78454, 2022.

\bibitem{shi2022csflow}
Hao Shi, Yifan Zhou, Kailun Yang, Xiaoting Yin, and Kaiwei Wang,
\newblock ``Csflow: Learning optical flow via cross strip correlation for
  autonomous driving,''
\newblock {\em IEEE Intelligent Vehicles Symposium}, pp. 1851--1858, 2022.

\bibitem{ahuja2019probabilistic}
Nilesh~A Ahuja, Ibrahima Ndiour, Trushant Kalyanpur, and Omesh Tickoo,
\newblock ``Probabilistic modeling of deep features for out-of-distribution and
  adversarial detection,''
\newblock {\em arXiv preprint arXiv:1909.11786}, 2019.

\bibitem{zavrtanik2021draem}
Vitjan Zavrtanik, Matej Kristan, and Danijel Sko{\v{c}}aj,
\newblock ``Draem-a discriminatively trained reconstruction embedding for
  surface anomaly detection,''
\newblock {\em IEEE International Conference on Computer Vision}, pp.
  8330--8339, 2021.

\bibitem{batzner2023efficientad}
Kilian Batzner, Lars Heckler, and Rebecca K{\"o}nig,
\newblock ``Efficientad: Accurate visual anomaly detection at millisecond-level
  latencies,''
\newblock {\em arXiv preprint arXiv:2303.14535}, 2023.

\bibitem{yu2021fastflow}
Jiawei Yu, Ye~Zheng, Xiang Wang, Wei Li, Yushuang Wu, Rui Zhao, and Liwei Wu,
\newblock ``Fastflow: Unsupervised anomaly detection and localization via 2d
  normalizing flows,''
\newblock {\em arXiv preprint arXiv:2111.07677}, 2021.

\bibitem{defard2021padim}
Thomas Defard, Aleksandr Setkov, Angelique Loesch, and Romaric Audigier,
\newblock ``Padim: a patch distribution modeling framework for anomaly
  detection and localization,''
\newblock {\em Pattern Recognition. ICPR International Workshops and
  Challenges: Virtual Event}, pp. 475--489, 2021.

\bibitem{roth2022towards}
Karsten Roth, Latha Pemula, Joaquin Zepeda, Bernhard Sch{\"o}lkopf, Thomas
  Brox, and Peter Gehler,
\newblock ``Towards total recall in industrial anomaly detection,''
\newblock {\em IEEE Conference on Computer Vision and Pattern Recognition}, pp.
  14318--14328, 2022.

\bibitem{deng2022anomaly}
Hanqiu Deng and Xingyu Li,
\newblock ``Anomaly detection via reverse distillation from one-class
  embedding,''
\newblock {\em IEEE Conference on Computer Vision and Pattern Recognition}, pp.
  9737--9746, 2022.

\bibitem{wang2021student}
Guodong Wang, Shumin Han, Errui Ding, and Di~Huang,
\newblock ``Student-teacher feature pyramid matching for anomaly detection,''
\newblock {\em arXiv preprint arXiv:2103.04257}, 2021.

\bibitem{rafiei2023computer}
Mehdi Rafiei, Jenni Raitoharju, and Alexandros Iosifidis,
\newblock ``Computer vision on x-ray data in industrial production and security
  applications: A comprehensive survey,''
\newblock {\em IEEE Access}, vol. 11, pp. 2445--2477, 2023.

\end{thebibliography}

\end{document}